%% file: main.tex
\documentclass[12pt]{article}

 \usepackage{amsmath}

\usepackage{graphicx}
\usepackage{color}

\usepackage[left=2.5cm,right=2.5cm,top=2.5cm,bottom=2.5cm]{geometry}
\usepackage[round]{natbib}

\usepackage{float}          
\usepackage{url}

\usepackage{dirtytalk}      

\usepackage{amssymb}        
\usepackage{mathtools}      
\usepackage{tikz}

\usepackage{colortbl}       
\usepackage{xcolor}
\definecolor{anchors_color}{HTML}{2B7BBA}

\usepackage{subcaption}     

\input{macros.tex}          

\begin{document}


\begin{center}
{\Large
	{\sc Understanding Post-hoc Explainers: \\ The Case of Anchors}
}
\bigskip

 Gianluigi Lopardo $^{1}$, Fr\'ed\'eric Precioso $^{2}$, Damien Garreau $^{1}$ 
\bigskip

{\it
$^{1}$ Université Côte d'Azur, Inria, CNRS, LJAD, France \\
$^{2}$ Université Côte d'Azur, Inria, CNRS, I3S, France 
\\ \{glopardo, dgarreau, fprecioso\}@unice.fr
}
\end{center}
\bigskip


{\bf R\'esum\'e.} 
Dans de nombreux sc\'enarios, l'interpr\'etabilit\'e de mod\`eles d'apprentissage automatique est une t\^ache hautement n\'ecessaire mais difficile. 
Pour expliquer les prédictions individuelles de ces mod\`eles, des approches locales et mod\`eles-agnostiques ont \'et\'e propos\'ees. 
Cependant, le processus de g\'en\'eration des explications peut \^etre, pour un utilisateur, aussi myst\'erieux que la pr\'ediction \`a expliquer. 
De plus, les m\'ethodes d'interpr\'etabilit\'e manquent souvent de garanties th\'eoriques et leur comportement sur des mod\`eles simples est souvent inconnu. 
S'il est difficile, voire impossible, de garantir qu'un explicateur se comporte comme pr\'evu sur un mod\`ele de pointe, nous pouvons au moins nous assurer que tout fonctionne sur des mod\`eles simples et d\'ejà interpr\'etables. 
Dans cet article, nous pr\'esentons un'analyse th\'eorique de Anchors \citep{ribeiro2018anchors} : une m\'ethode populaire d'interpr\'etabilit\'e bas\'ee sur des r\`egles qui met en \'evidence un petit ensemble de mots pour expliquer la d\'ecision d'un classificateur de texte. 
Apr\`es avoir formalis\'e son algorithme et fourni des informations utiles, nous d\'emontrons math\'ematiquement que Anchors produisent des r\'esultats significatifs lorsqu'elles sont utilis\'ees avec des classificateurs de texte lin\'eaires en plus d'une vectorisation TF-IDF. 
Nous pensons que 
notre cadre d'analyse peut contribuer au d\'eveloppement de nouvelles m\'ethodes d'explicabilit\'e bas\'ees sur des fondements th\'eoriques solides.

{\bf Mots-cl\'es.} Apprentissage statistique, Classification supervisée et non supervisée, Interprétabilité, Traitement du Langage Naturel. 

\medskip

{\bf Abstract.}
%
In many scenarios, the interpretability of machine learning models is a highly required but difficult task. 
To explain the individual predictions of such models, local model-agnostic approaches have been proposed. 
However, the process generating the explanations can be, for a user, as mysterious as the prediction to be explained. 
Furthermore, interpretability methods frequently lack theoretical guarantees, and their behavior on simple models is frequently unknown. 
While it is difficult, if not impossible, to ensure that an explainer behaves as expected on a cutting-edge model, we can at least ensure that everything works on simple, already interpretable models.
In this paper, we present a theoretical analysis of Anchors \citep{ribeiro2018anchors}: a popular rule-based interpretability method that highlights a small set of words to explain a text classifier's decision. 
After formalizing its algorithm and providing useful insights, we demonstrate mathematically that Anchors produces meaningful results when used with linear text classifiers on top of a TF-IDF vectorization.
We believe that 
our analysis framework can aid in the development of new explainability methods based on solid theoretical foundations. 


\bigskip\bigskip


\begin{figure}[b]
\centering
\begin{minipage}{0.3\textwidth}\centering
    \includegraphics[scale=0.8]{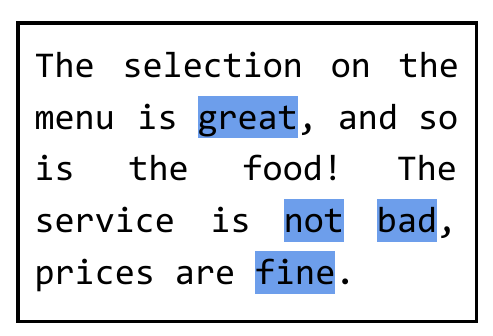}
\end{minipage}
\begin{minipage}{0.3\textwidth}\centering
    $\texttt{precision}: \; 0.97$ \\
    $\texttt{coverage}: \; 0.12$
\end{minipage}
    \caption{\label{fig:example}Anchors explaining the positive prediction of a black-box model~$f$ on an example~$\xi$ from the Restaurant review dataset. The anchor $A = \{\textit{great, not, bad, fine}\}$ (in blue), having length $\length{A} = 4$ is selected. Intuitively, that a document contains these four words together ensures a positive prediction by $f$ with high probability ($\texttt{precision} = \; 0.97$), while being not too uncommon ($\texttt{coverage} = \; 0.12$).}
\end{figure}
\section{Introduction}
\label{sec:introduction}
Complex machine learning models with billions of parameters, such BERT and GPT-3 \citep{devlin2019bert,brown_et_al_2020}, have become increasingly popular in natural language processing.
The interpretability of these models, however, continues to be a problem in sensitive or important scenarios when consumers and subject-matter experts demand explanations.
Many solutions, including local model-agnostic approaches that explicate individual predictions for a particular instance, have been proposed for creating interpretable explanations to overcome this issue.
These methods, however, may lack theoretical guarantees, and their behavior on simple, interpretable models is frequently unknown, potentially leading to misleading results. 

In this work, we focus on Anchors \citep{ribeiro2018anchors}, an increasingly popular local model-agnostic, and we particularly deal with its implementation for text data. 
In this context, Anchors outputs a list of words that, if present, produce the same prediction with a high probability, and are presented to the user as such (see Figure~\ref{fig:example}). 
We examine whether Anchors, especially when the model being explained is already interpretable, can pinpoint the most crucial words for the prediction. 
In this paper, we present the framework for analyzing Anchors on linear text classifiers proposed in \citet{lopardo2022sea}. 
Our analysis lays the groundwork for future research in the theoretical foundations of interpretability and offers insightful information about the behavior of Anchors for text data. 
\paragraph{Paper organization.}
The structure of the paper is as follows. 
We begin by introducing some related works on interpretability and its theoretical foundations in the paragraph after. 
Second, we formalize Anchors' text classification mechanism in Section \ref{sec:anchors} and explain its fundamental concepts. 
We then go over the definition of a more tractable, exhaustive version of the algorithm in Section~\ref{sec:exhaustive-anchors}, which is the main focus of our research. 
Next, in Section~\ref{sec:analysis}, we illustrate a theoretical and empirical analysis of Anchors' behavior on linear classifiers to better understand their efficacy for text data. 
Finally, we summarize the results of our study in Section~\ref{sec:conclusion} and reach our conclusions.

\paragraph{Related work.}
\label{sec:related}
In recent years, several methods have been proposed for machine learning interpretability \citep{guidotti2018local, adadi2018peeking,linardatos2021explainable}, and among them, rule-based methods have gained popularity. 
This is because users prefer rule-based explanations \citep{lim2009and, stumpf2007toward}, and hierarchical decision lists \citep{wang2015falling} can be useful for understanding the global behavior of a model. 
However, smaller and disjoint rules are easier to interpret, and \citet{lakkaraju2016interpretable} introduces the concept of \emph{coverage} to extract small and disjoint rules while compromising between accuracy and interpretability. Alternatively, \citet{barbiero2022entropy} proposes to learn simple logical rules along with the parameters of the model itself, so as not to sacrifice accuracy.

Several approaches have also focused on \emph{local interpretability}, where, typically, a simpler model is used to approximate any black-box model around a specific instance to explain. 
For example, LORE \citep{guidotti2018local} uses a decision tree as a local surrogate, while LIME \citep{ribeiro2016should} and SHAP \citep{lundberg2017unified} provide explanations using a linear model trained on perturbed samples of the instance to explain. 
\citet{amoukou2021consistent} proposed Minimal Sufficient Rules, similar to Anchors for tabular data, extended to regression models, which can directly deal with continuous features with no need for discretization.
While LIME and SHAP assign a weight to each word of the example, Anchors extracts the minimal subset of words that is sufficient to have the same prediction as the example in high probability.

Despite the popularity of interpretability methods, few works have investigated their theoretical guarantees. 
For feature importance methods, \citet{lundberg2017unified} provides insights into the case of linear models for kernel SHAP, while \citet{mardaoui_garreau_2021} extended \citet{garreau2020explaining} and investigated LIME for text data. 
In this paper, we exhibit an analysis of Anchors on linear text classifiers, first presented in \citet{lopardo2022sea}. 
Having theoretical guarantees makes it possible not only to act as a sanity check for a method, but also to compare different explainers with each other, which is otherwise only possible experimentally \citep{lopardo2022comparing}. 



\section{Anchors for text data}
\label{sec:anchors}
This section describes Anchors for text data, as introduced by \citet{ribeiro2018anchors}. 
We start by defining the setting and notation, followed by presenting the key concepts of precision and coverage.
Finally, we explain the algorithm and discuss the sampling scheme. 

\paragraph{Setting and notation.}
\label{sec:setting-notation}
The problem at hand is to explain the decision of a binary classifier~$f$ that takes documents as input. 
We denote a generic document by $\doc$ and the particular example being explained by Anchors as $\xi$. 
We define a global dictionary $\dictionary$ of cardinality $D$ and represent any document as a finite sequence of elements from $\dictionary$. 
We also define a local dictionary $\localdic_\xi$ for a given document $\xi$ as the set of distinct words in $\xi$. 

We make two restrictive assumptions about the class of models they consider. 
First, we restrict the analysis to binary classification. 
Second, we assume that the classifier $g$ relies on a vectorization of the documents, where $g = h \circ \Vectorizer$, and $\Vectorizer$ is a deterministic mapping from texts to $\Reals^D$, and $h : \Reals^D \to \Reals^p$ is a given measurable function.

We define an anchor as any non-empty subset of $[b]$, corresponding to a preserved set of words of $\xi$, and denote the set of all candidate anchors for $\xi$ as $\Anchors$. For any anchor $A \in \Anchors$, they set $\length{A}$ as the length of the anchor, defined as the number of words contained in $A$. In practice, an anchor $A$ for a document $\xi$ is represented as a non-empty sublist of the words present in the document.

\paragraph{Precision and coverage.}
\label{sec:precision-coverage}
The precision of an anchor $A$ is defined as the probability for a local perturbation of $\xi$ to be classified as $1$, and it can be written as $\precision{A} = \probaunder{g(\tfidf{x}) = 1}{A}$, where $x$ is a random perturbation of $\xi$ still containing all the words included in $A$. 
The coverage measures the proportion of examples in a corpus that satisfy a given anchor, \emph{i.e.} $\coverage{A} = \probaunder{A}{f(x) = 1}$, where the expectation is taken with respect to $x$, a random perturbation of $\xi$ still containing all the words included in $A$. 
Anchors algorithm prioritizes the precision and coverage: in a nutshell, \textbf{Anchors will search for an anchor of maximum coverage with prescribed precision.} 
In the section after, we go into more detail.

\paragraph{Algorithm.}
\label{sec:algoritgm}
In practice, computing the coverage can be expensive and may not be feasible if a corpus is unavailable during prediction. 
To address this issue, the default implementation minimizes the length of anchors instead of maximizing the coverage, as shorter anchors tend to have larger coverage. 
The optimization problem in this case is to find an anchor with minimal length $A \in \Anchors$ that satisfy $\precision{A} \geq 1 - \epsilon$, where $\epsilon$ is a tolerance threshold (set to $0.05$ in practice). 
The precision of a specific anchor cannot be computed exactly because the expectation in the precision formula cannot be calculated in general. 
Instead, the precision is approximated using $\Empprec_n(A)$, an empirical approximation defined in Section~\ref{sec:exhaustive-anchors}. 
Since the cardinality of $\Anchors$ is too large in practical scenarios, the default implementation applies the KL-LUCB algorithm to identify a subset of high-precision rules that serve as representatives of all candidate anchors to approximate the solution to the optimization problem. 
However, this paper focuses on an exhaustive version of Anchors described in Section~\ref{sec:exhaustive-anchors}, and does not consider the optimization procedure.

\paragraph{Sampling.}
\label{sec:sampling}
The objective is to observe the behavior of the model $f$ in a local neighborhood of $\xi$, while holding the set of words in $A$ fixed. 
To achieve this, Anchors generates perturbed samples $x_1,\ldots,x_n$, for a given example $\xi$ and a candidate anchor $A \in \Anchors$. 
This process involves creating $n$ identical copies of the example, drawing a number of copies to be perturbed for each word not in the anchor, and then replacing the selected words belonging to those copies with the token \emph{UNK.} 
Note that this is done in practice in the official implementation, but it is different from the perturbation distribution used in \cite{ribeiro2018anchors}, where selected words are replaced with others having the same part-of-speech tag with probability proportional to their similarity in an embedding space.

Replacing words with a predefined token can generate meaningless sentences that can deceive a classifier and produce unrealistic samples \citep{hase2021out}. 
Nevertheless, we consider the \emph{UNK}-replacement in this work because it is the default choice used by Anchors' users and replicates word removals exactly in the case of TF-IDF vectorization. 
Additionally, our experiments show that the results remain valid when words are replaced using \emph{BERT}, which is an alternative solution propose din Anchors' package. 
Anchors' sampling process is similar to LIME for text data \citep{mardaoui_garreau_2021}, except LIME removes all occurrences of a given word when it is selected for removal. 
The sampling process can be simplified as follows: for any sample $x_i$, each word $x_{i,k}$ such that $k\notin A$ is replaced independently with probability $1/2$. 
\citet{lopardo2022sea} shows that for any given anchor $A$, the random variable $\Mult_j$, which is defined as the multiplicity of word $\word_j$ in the perturbed sample $x$, can be described by $\Mult_j \sim \anchor_j + \binomial{\mult_j-\anchor_j}{1/2}$, where $\anchor_j$ is the number of occurrences of $\word_j$ in $A$.


\section{Exhaustive $p$-Anchors}
\label{sec:exhaustive-anchors}
In this section, we introduce exhaustive $p$-Anchors as the central object of our study. 
This is a formalized version of the original combinatorial optimization problem presented in Section \ref{sec:anchors}, which can be solved for any evaluation function $p:\Anchors\to\Reals$. 

\paragraph{Description of the algorithm.}
\label{sec:exhaustive-description}
The optimization problem of Anchors can be divided into two steps. 
Firstly, we select all anchors in $\Anchors$ such that $\precision{A}\geq 1-\epsilon$ to obtain a subset of anchors $\Anchors_1(\epsilon)$. 
The full anchor $[b]$ has precision $1$ and is always included in this set. 
Secondly, we select the anchors in $\Anchors_1(\epsilon)$ that have minimal length to obtain a subset of anchors $\Anchors_2(\epsilon)$. 
Finally, we select the anchor(s) with the highest precision from $\Anchors_2(\epsilon)$, which we call $\Anchors_3(\epsilon)$. 
If $\Anchors_3(\epsilon)$ contains more than one anchor, we randomly select one. 

We have designed this algorithm to be flexible so that it can be used with different evaluation functions. For example, we can use the algorithm with $p=\Empprec_n$ or $p=\Precision$ as a selection function, or any other function that approximates $\Precision$ well. When we use $p=\Precision$, we refer to this version of the algorithm as \emph{exhaustive Anchors}. When we use $p=\Empprec_n$, we refer to this version as \emph{empirical Anchors}.

Empirical Anchors is very similar to Anchors, but instead of using an efficient approximate procedure, it considers all possible anchors. 
The main difference is that empirical Anchors selects anchors with the maximal precision in the third step, while this is not necessarily the case with the default implementation. 
We refer to \citet{lopardo2022sea} for a more detailed description, where \emph{empirical Anchors} and \emph{exhaustive Anchors} are mathematically and empirically proved to be close.


\section{Analysis on Linear Classifiers}
\label{sec:analysis}
In this Section we detail the framework used in \citet{lopardo2022sea} to analyze Anchors on linear classifiers. 
First, we introduce the vectorizer that we are considering, and then delve into the analysis of Anchors' behavior on linear models, presenting our key findings. 
The interested reader can find mathematical proofs and empirical validations of all our claims in the Appendix of \citet{lopardo2022sea}. 

\paragraph{Vectorizers.}
\label{sec:vectorizers}
Natural language processing classifiers rely heavily on a vector representation $\Vectorizer$ of documents, which is often obtained using the popular TF-IDF (Term Frequency-Inverse Document Frequency) transform \citep{luhn1957statistical}. 
This method assigns greater weight to words that appear frequently in a particular document $\doc$, but less frequently in the overall corpus $\corpus$. 
In this paper, we assume that models work with a non-normalized TF-IDF vectorizer, which is defined by a vector $\tfidf{\doc}$ that is based on the inverse document frequency (IDF) of each word in the vocabulary. 
Once the TF-IDF vectorizer is fitted to a corpus, the vocabulary is fixed, and any word not present in the initial corpus is assigned an IDF term of zero.

Note that with this vectorizer, replacing any word with a fixed token \emph{UNK} is equivalent to simply removing it, as the token is not present in the initial corpus. 
We also remark that when models are trained on a (non-normalized) TF-IDF vectorization, the exact location of the words in the document does not matter, and thus only the occurrences of each word in an anchor $A$ are important when computing precision. 
We represent an anchor $A = (\anchor_1,\ldots,\anchor_d)$, where $\anchor_j \leq \mult_j$ for all $j \in [D]$ and $\anchor_j = 0$ for any $j > d$. 
Finally, note that the TF-IDF of $\word_j$ in the perturbed sample can be expressed as $(\anchor_j+\binomial{\mult_j-\anchor_j}{1/2})\idf_j$, which intuitively corresponds to the number of occurrences of $\word_j$ in the anchor plus a random number of occurrences depending on the sampling. 

\paragraph{Linear classifiers.}
\label{sec:linear}
We now focus on linear classifiers in this section. For any document $\doc$, we define the linear classifier $f(\doc)$ as follows:
\begin{equation}
\label{eq:def-linear-classifier}
f(\doc) = \indic{\lambda^\top \tfidf{\doc} + \lambda_0 > 0}
\, ,
\end{equation}
where $\lambda \in \Reals^D$ is a vector of coefficients, and $\lambda_0 \in \Reals$ is an intercept. 
Equation \eqref{eq:def-linear-classifier} includes several examples, two of which are the perceptron \citep{rosenblatt_1958}, which predicts exactly as in Equation \eqref{eq:def-linear-classifier}, and logistic models that predict as $1$ if $\sigma(\lambda^\top \tfidf{\doc} + \lambda_0) > 1/2$. 
Here, $\sigma : \Reals \to \sigma(t) = \frac{1}{1 + \exps{-t}} \in [0,1]$ is the logistic function. 
As $\sigma(y)>1/2$ if, and only if, $y>0$, logistic models can also be rewritten as in Equation \eqref{eq:def-linear-classifier}. 
For a more complete overview of linear classifiers, refer to Chapter 4 in \citet{hastie_et_al_2009}.
We investigate the precision of a linear classifier using a Berry-Esseen-type statement \citep{berry_1941,esseen_1942}. 
\begin{proposition}[Precision of a linear classifier]
\label{prop:precision-logistic}
Let $\lambda,\lambda_0$ be the coefficients associated to the linear classifier defined by Eq.~\eqref{eq:def-linear-classifier}. 
Assume that for all $j\in [d]$, $\lambda_j\idf_j\neq 0$. 
Define, for all $A \in\Anchors$, 
\begin{align}
\label{eq:arg-approx-prec-logistic}
\approxprec{A} & \defeq \frac{- \lambda_0 - \frac{1}{2} \sum_{j=1}^d \lambda_j \idf_j (\mult_j + \anchor_j)}{\sqrt{\frac{1}{4} \sum_{j=1}^d \lambda_j^2 \idf_j^2 (\mult_j - \anchor_j)}}
\, .
\end{align}
Let $\Phibar\defeq 1-\Phi$, where $\Phi$ denotes the cumulative distribution function of a $\gaussian{0}{1}$. 
Then, for any $A\in\Anchors$ such that $\length{A} \leq b / 2$, 
\begin{align}
\label{eq:precision-logistic}
\abs{\precision{A} - \phibar{\approxprec{A}}} & \leq \\ \nonumber
C\cdot \left(\frac{\max_j \lambda_j^2\idf_j^2}{\min_j \lambda_j^2\idf_j^2}\right)^{3/2}  & \cdot \left(\frac{\max_j\mult_j}{\min_j\mult_j}\right)^{3/2} \cdot \frac{1}{\sqrt{d}}
\, ,
\end{align}
where $C\approx 7.15$ is a numerical constant. 
\end{proposition}

In practice, the precision of the anchor can be well approximated by $\Phibar\circ \Approxprec$, 
where, $\Phibar\defeq 1-\Phi$, and $\Phi$ denotes the cumulative distribution function of a $\gaussian{0}{1}$. 
We can use this approach to study exhaustive $p$-Anchors with $p=\Phibar \circ \Approxprec$ instead of exhaustive Anchors. 
Proposition \ref{prop:precision-logistic} is proven in 
\citet{lopardo2022sea}, where it is shown that in the case of normalized TF-IDF, a constant with the same rate appears. 

A typical value for $\idf_j$ and $\mult_j$ in our setting lies between $1$ and $10$. 
Thus, the main assumption is the absence of zero components in $\lambda$. 
The fact that an explanation based on more than half the document is factually uninterpretable justifies the assumption regarding the length of anchors (less than half the length of the document).
The constant would also be improved by assuming even smaller anchors. 

We can now focus on exhaustive $\Phibar\circ\Approxprec$-Anchors. 
Let us set $\gamma \defeq \lambda_0 + \sum_j \lambda_j\idf_j\mult_j$. 
Note that, since we assume $f(\xi)=1$, $\gamma >0$. 
Let us also set
\begin{equation}
\label{eq:def-positive-anchors}
\Anchors_+ \defeq \{A\in\Anchors, \text{ s.t. } \anchor_j > 0 \Rightarrow \lambda_j > 0\}
\, ,
\end{equation}
the set of anchors with support corresponding to words with a positive influence. 
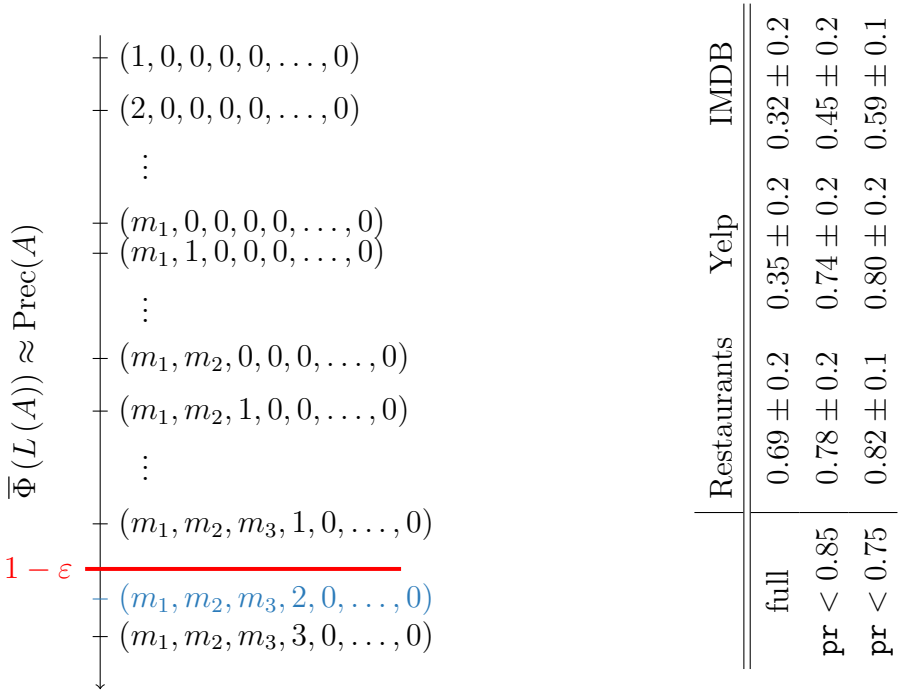
\begin{figure}[t]
 \centering
    \begin{subfigure}[b]{0.45\textwidth}
    \centering
        \input{figs/linear.tex}
    \end{subfigure}
    \begin{subfigure}[b]{0.45\textwidth}
    \centering
       \rotatebox{90}{
        \input{figs/linear_tfidf}

        }
    \end{subfigure}
    \caption{\label{fig:linear} On the left, illustration of Proposition \ref{prop:approx-prec-maximization}. On linear models, the algorithm includes words having the highest $\lambda_j\idf_j$s first. Finally, the minimal anchor satisfying the precision condition $\phibar{\approxprec{A}} \approx\precision{A} \geq 1-\varepsilon$ is selected, which is $A=(\mult_1,\mult_2,\mult_3,2,0,\ldots,0,0)$ in the example. On the right, validation of Proposition~\ref{prop:approx-prec-maximization}. Average Jaccard similarity between the anchor $A$ and the first $\length{A}$ words ranked by $\lambda_j\idf_j$ for a logistic model on positive documents and low-confidently classified subset ($\texttt{pr} = g(\tfidf{\xi}) < 0.85$, or $\texttt{pr} < 0.75$).}
\end{figure}
\begin{proposition}[Approximate precision maximization]
\label{prop:approx-prec-maximization}
Assume the words to be ordered such that $\lambda_1\idf_1 > \lambda_2\idf_2 > \cdots > \lambda_d\idf_d$, with at least one $\lambda_j$ greater than zero. 
Assume that $\lambda_0 > -\gamma / 2$. 
Then the Algorithm 
applied to the selection function $p\defeq\Phibar \circ L$ will select an anchor $A^p\in\Anchors_+$ such that there exists $j_0\in [d]$ with the following property: for all $j<j_0$, $\anchor_j=\mult_j$, $\anchor_{j_0}\leq\mult_j$, and for all $j\geq j_0$, $\anchor_j=0$. 
\end{proposition}

In other words, Proposition~\ref{prop:approx-prec-maximization} implies that Anchors keeps only words that have a positive influence on the prediction for a linear classifier: this is a reassuring property. 
Furthermore, it prioritizes the words with the highest $\lambda_j\idf_j$s, adding them to the anchor until the precision condition is met. 
See Figure \ref{fig:linear} for an illustration. 

We conducted the following experiment to empirically validate this Property. 
We first trained a logistic model on three review datasets, achieving accuracies between $85\%$ and $88\%$ on the test set. 
We then ran Anchors with the default setting $10$ times on positively classified documents. 
For each document, we measure the Jaccard similarity between the anchor $A$ and the first $\length{A}$ words ranked by $\lambda_j\idf_j$.  
Right panel on Figure~\ref{fig:linear} reports the average Jaccard index: results confirm  Proposition~\ref{prop:approx-prec-maximization}. 

It is important to note that the official implementation of Anchors is not designed to output the best anchor. 
In cases where the prediction is \emph{easy}, such as when $g(\tfidf{\xi}) \geq 0.75$ or $g(\tfidf{\xi}) \geq 0.85$, there may be multiple anchors that satisfy the criteria, and the algorithm selects one at random. 
This explains the varying similarity between the full dataset and the harder subsets. 


\section{Conclusion}
\label{sec:conclusion}
In this paper, we presented the first theoretical analysis of Anchors, focusing on its implementation for textual data and providing insights on the sampling procedure. 
Our study mainly focused on Anchors' behavior on linear models, and to this end, we introduced an approximate, tractable version of the algorithm that is similar to the default implementation. 
Our analysis demonstrated that Anchors provides meaningful results when applied to these models, which is supported by experiments with the official implementation. 

Our work highlights the significance of theoretical analysis in the development of explainability methods. 
We believe that the insights presented in this paper will be valuable for researchers and practitioners in natural language processing in correctly comprehending the explanations provided by Anchors. 
Moreover, the analytical framework we presented can benefit the explainability community by facilitating the development of new methods founded on robust theoretical principles and analyzing current ones. 
As future work, we plan to extend this analysis to other classes of models and different types of data, including images and tabular data. 

\paragraph{Acknowledgements.}
This work has been supported by the French government, through the NIM-ML project (ANR-21-CE23-0005-01), and by EU Horizon 2020 project AI4Media (contract no.~951911). 





\bibliography{references}
\end{document}

%% file: macros.tex
\newtheorem{proposition}{Proposition}

\newcommand{\indic}[1]{\mathbf{1}_{#1}}                 
\newcommand{\Reals}{\mathbb{R}}                         
\newcommand{\probaunder}[2]{\mathbb{P}_{#2}\left (#1\right )}   
\newcommand{\binomial}[2]{\mathrm{Bin}(#1,#2)}          
\newcommand{\exps}[1]{\mathrm{e}^{#1}}                  
\newcommand{\Phibar}{\overline{\Phi}}                           
\newcommand{\phibar}[1]{\Phibar\left(#1\right)}                 
\newcommand{\gaussian}[2]{\mathcal{N}(#1,#2)}           
\newcommand{\defeq}{\vcentcolon =}                      
\renewcommand{\epsilon}{\varepsilon}
\newcommand{\abs}[1]{\left\lvert#1\right\rvert}         

\newcommand{\length}[1]{\left\lvert#1\right\rvert}              
\newcommand{\doc}{z}                                            
\newcommand{\word}{w}                                           
\newcommand{\dictionary}{\mathcal{D}}                           
\newcommand{\Vectorizer}{\varphi}                               
\newcommand{\Anchors}{\mathcal{A}}                              
\newcommand{\localdic}{\mathcal{D}}                             
\newcommand{\Precision}{\mathrm{Prec}}                          
\newcommand{\precision}[1]{\Precision(#1)}                      
\newcommand{\idf}{v}                                            
\newcommand{\Tfidf}{\varphi}
\newcommand{\tfidf}[1]{\Tfidf(#1)}                              
\newcommand{\coverage}[1]{\mathrm{Cov}(#1)}                     
\newcommand{\Empprec}{\widehat{\mathrm{Prec}}}                  
\newcommand{\mult}{m}                                           
\newcommand{\Mult}{M}                                           
\newcommand{\anchor}{a}                                         
\newcommand{\corpus}{\mathcal{C}}                               
\newcommand{\Approxprec}{L}                                     
\newcommand{\approxprec}[1]{\Approxprec\left(#1\right)}         

%% file: figs/linear.tex
\begin{tikzpicture}
    \draw[<-] (0,-0.7) -- (0,8) node[left, midway, rotate=90, yshift=1.0cm, xshift=2cm] {$\phibar{\approxprec{A}} \approx\precision{A}$};

    \draw (-0.1,7.7) -- (0.1,7.7) node [right] {$(1,0,0,0,0,\ldots,0)$};
    \draw (-0.1,7.0) -- (0.1,7.0) node [right] {$(2,0,0,0,0,\ldots,0)$};
    \draw (0.0, 6.35) node [right, xshift=0.4cm] {$\vdots$};
    \draw (-0.1,5.5) -- (0.1,5.5) node [right] {$(\mult_1,0,0,0,0,\ldots,0)$};
    \draw (-0.1,5.1) -- (0.1,5.1) node [right] {$(\mult_1,1,0,0,0,\ldots,0)$};
    \draw (0.0, 4.45) node [right, xshift=0.4cm] {$\vdots$};
    \draw (-0.1,3.7) -- (0.1,3.7) node [right] {$(\mult_1,\mult_2,0,0,0,\ldots,0)$};
    \draw (-0.1,3.0) -- (0.1,3.0) node [right] {$(\mult_1,\mult_2,1,0,0,\ldots,0)$};
    \draw (0.0, 2.35) node [right, xshift=0.4cm] {$\vdots$};
    \draw (-0.1,1.5) -- (0.1,1.5) node [right] {$(\mult_1,\mult_2,\mult_3,1,0,\ldots,0)$};
    \draw[anchors_color] (-0.1,0.5) -- (0.1,0.5) node [right] {$(\mult_1,\mult_2,\mult_3,2,0,\ldots,0)$};
    \draw (-0.1,0.0) -- (0.1,0.0) node [right] {$(\mult_1,\mult_2,\mult_3,3,0,\ldots,0)$};

    \draw[red, line width=1.5pt] (-0.2,0.9) node [left] {$1-\varepsilon$} -- (4.0,0.9);

\end{tikzpicture}

%% file: figs/linear_tfidf.tex
     %
    \setlength{\extrarowheight}{4pt}%
    \setlength{\tabcolsep}{5pt}
    \begin{tabular}{c | c c c}
     & Restaurants & Yelp & IMDB \\
    \hline\hline
    full & $0.69 \pm 0.2$ & $0.35 \pm 0.2$ & $0.32 \pm 0.2$ \\
    $\texttt{pr} < 0.85$ & $0.78 \pm 0.2$ & $0.74 \pm 0.2$ & $0.45 \pm 0.2$ \\
    $\texttt{pr} < 0.75$ & $0.82 \pm 0.1$ & $0.80 \pm 0.2$ & $0.59 \pm 0.1$ \\ 
    \end{tabular}

%% file: main.bbl
\begin{thebibliography}{}

\bibitem[Adadi and Berrada, 2018]{adadi2018peeking}
Adadi, A. and Berrada, M. (2018).
\newblock Peeking inside the black-box: a survey on explainable artificial
  intelligence ({XAI}).
\newblock {\em IEEE access}, 6:52138--52160.

\bibitem[Amoukou and Brunel, 2022]{amoukou2021consistent}
Amoukou, S.~I. and Brunel, N. J.~B. (2022).
\newblock {Consistent Sufficient Explanations and Minimal Local Rules for
  explaining regression and classification models}.
\newblock In {\em Advances in Neural Information Processing Systems}.

\bibitem[Barbiero et~al., 2022]{barbiero2022entropy}
Barbiero, P., Ciravegna, G., Giannini, F., Li{\'o}, P., Gori, M., and Melacci,
  S. (2022).
\newblock Entropy-based logic explanations of neural networks.
\newblock In {\em Proceedings of the AAAI Conference on Artificial
  Intelligence}, volume~36, pages 6046--6054.

\bibitem[Berry, 1941]{berry_1941}
Berry, A.~C. (1941).
\newblock The accuracy of the gaussian approximation to the sum of independent
  variates.
\newblock {\em {Transactions of the American Mathematical Society}},
  49(1):122--136.

\bibitem[Brown et~al., 2020]{brown_et_al_2020}
Brown, T., Mann, B., Ryder, N., Subbiah, M., Kaplan, J.~D., Dhariwal, P.,
  Neelakantan, A., Shyam, P., Sastry, G., Askell, A., et~al. (2020).
\newblock Language models are few-shot learners.
\newblock {\em {Advances in Neural Information Processing Systems}},
  33:1877--1901.

\bibitem[Devlin et~al., 2019]{devlin2019bert}
Devlin, J., Chang, M.-W., Lee, K., and Toutanova, K. (2019).
\newblock {{BERT}: Pre-training of Deep Bidirectional Transformers for Language
  Understanding}.
\newblock In {\em Proceedings of the 2019 Conference of the North {A}merican
  Chapter of the Association for Computational Linguistics: Human Language
  Technologies, Volume 1 (Long and Short Papers)}, pages 4171--4186,
  Minneapolis, Minnesota. Association for Computational Linguistics.

\bibitem[Esseen, 1942]{esseen_1942}
Esseen, C.-G. (1942).
\newblock {On the Liapunov limit error in the theory of probability}.
\newblock {\em Ark. Mat. Astr. Fys.}, 28:1--19.

\bibitem[Garreau and Luxburg, 2020]{garreau2020explaining}
Garreau, D. and Luxburg, U. (2020).
\newblock {Explaining the explainer: A first theoretical analysis of {LIME}}.
\newblock In {\em International Conference on Artificial Intelligence and
  Statistics}, pages 1287--1296. PMLR.

\bibitem[Guidotti et~al., 2018]{guidotti2018local}
Guidotti, R., Monreale, A., Ruggieri, S., Pedreschi, D., Turini, F., and
  Giannotti, F. (2018).
\newblock Local rule-based explanations of black box decision systems.
\newblock {\em arXiv preprint arXiv:1805.10820}.

\bibitem[Hase et~al., 2021]{hase2021out}
Hase, P., Xie, H., and Bansal, M. (2021).
\newblock The out-of-distribution problem in explainability and search methods
  for feature importance explanations.
\newblock {\em Advances in Neural Information Processing Systems},
  34:3650--3666.

\bibitem[Hastie et~al., 2009]{hastie_et_al_2009}
Hastie, T., Tibshirani, R., and Friedman, J.~H. (2009).
\newblock {\em The elements of statistical learning: data mining, inference,
  and prediction}, volume~2.
\newblock Springer.

\bibitem[Lakkaraju et~al., 2016]{lakkaraju2016interpretable}
Lakkaraju, H., Bach, S.~H., and Leskovec, J. (2016).
\newblock Interpretable decision sets: A joint framework for description and
  prediction.
\newblock In {\em Proceedings of the 22nd ACM SIGKDD international conference
  on knowledge discovery and data mining}, pages 1675--1684.

\bibitem[Lim et~al., 2009]{lim2009and}
Lim, B.~Y., Dey, A.~K., and Avrahami, D. (2009).
\newblock Why and why not explanations improve the intelligibility of
  context-aware intelligent systems.
\newblock In {\em Proceedings of the SIGCHI conference on human factors in
  computing systems}, pages 2119--2128.

\bibitem[Linardatos et~al., 2021]{linardatos2021explainable}
Linardatos, P., Papastefanopoulos, V., and Kotsiantis, S. (2021).
\newblock {Explainable AI: A Review of Machine Learning Interpretability
  Methods}.
\newblock {\em Entropy}, 23(1):18.

\bibitem[Lopardo and Garreau, 2022]{lopardo2022comparing}
Lopardo, G. and Garreau, D. (2022).
\newblock {Comparing Feature Importance and Rule Extraction for
  Interpretability on Text Data}.
\newblock In {\em ICPR 2-nd Workshop on Explainable and Ethical AI - 26TH
  International Conference on Pattern Recognition (XAIE @ ICPR) 2022}.

\bibitem[Lopardo et~al., 2023]{lopardo2022sea}
Lopardo, G., Precioso, F., and Garreau, D. (2023).
\newblock {A Sea of Words: An In-Depth Analysis of Anchors for Text Data}.
\newblock In {\em International Conference on Artificial Intelligence and
  Statistics}.

\bibitem[Luhn, 1957]{luhn1957statistical}
Luhn, H.~P. (1957).
\newblock A statistical approach to mechanized encoding and searching of
  literary information.
\newblock {\em IBM Journal of research and development}, 1(4):309--317.

\bibitem[Lundberg and Lee, 2017]{lundberg2017unified}
Lundberg, S.~M. and Lee, S.-I. (2017).
\newblock {A Unified Approach to Interpreting Model Predictions}.
\newblock {\em Advances in Neural Information Processing Systems},
  30:4765--4774.

\bibitem[Mardaoui and Garreau, 2021]{mardaoui_garreau_2021}
Mardaoui, D. and Garreau, D. (2021).
\newblock An analysis of {LIME} for text data.
\newblock In {\em International Conference on Artificial Intelligence and
  Statistics}, pages 3493--3501. PMLR.

\bibitem[Ribeiro et~al., 2016]{ribeiro2016should}
Ribeiro, M.~T., Singh, S., and Guestrin, C. (2016).
\newblock {\say{Why should I trust you?} Explaining the predictions of any
  classifier}.
\newblock In {\em Proceedings of the 22nd ACM SIGKDD international conference
  on knowledge discovery and data mining}, pages 1135--1144.

\bibitem[Ribeiro et~al., 2018]{ribeiro2018anchors}
Ribeiro, M.~T., Singh, S., and Guestrin, C. (2018).
\newblock Anchors: High-precision model-agnostic explanations.
\newblock In {\em Proceedings of the {AAAI} Conference on Artificial
  Intelligence}, volume~32.

\bibitem[Rosenblatt, 1958]{rosenblatt_1958}
Rosenblatt, F. (1958).
\newblock The perceptron: a probabilistic model for information storage and
  organization in the brain.
\newblock {\em Psychological review}, 65(6):386.

\bibitem[Stumpf et~al., 2007]{stumpf2007toward}
Stumpf, S., Rajaram, V., Li, L., Burnett, M., Dietterich, T., Sullivan, E.,
  Drummond, R., and Herlocker, J. (2007).
\newblock Toward harnessing user feedback for machine learning.
\newblock In {\em Proceedings of the 12th international conference on
  Intelligent user interfaces}, pages 82--91.

\bibitem[Wang and Rudin, 2015]{wang2015falling}
Wang, F. and Rudin, C. (2015).
\newblock Falling rule lists.
\newblock In {\em Artificial intelligence and statistics}, pages 1013--1022.
  PMLR.

\end{thebibliography}
